\documentclass{article}
\usepackage{jfrExamplee}
\usepackage{graphicx}
\usepackage{setspace}
\usepackage{algorithm}
\usepackage{algpseudocode}
\usepackage{amsmath}
\usepackage{amssymb}
\usepackage{xcolor}
\usepackage{subfig}
\usepackage{natbib}
\usepackage{authblk}
\usepackage{url}
\usepackage{ragged2e}
\usepackage{lineno}

\title{Data-Driven Worker Activity Recognition and Efficiency Estimation in Manual Fruit Harvesting}

\author{
\textbf{Uddhav Bhattarai}$^{1,*}$, \textbf{Rajkishan Arikapudi}$^1$, \textbf{Steven A. Fennimore}$^2$, \textbf{Frank N Martin}$^3$, \textbf{Stavros G. Vougioukas}$^{1,*}$}

\small{
\affil{$^1$Department of Biological and Agricultural Engineering, University of California, Davis, CA, 95616, USA}
\affil{$^2$Department of Plant Sciences, University of California, Davis, CA, 95616, USA}
\affil{$^3$Crop Improvement and Protection Research Unit, U.S. Department of Agriculture Agricultural Research Service, Salinas, CA 93905, USA}
\affil{$^*$Corresponding authors: ubhattarai@ucdavis.edu, svougioukas@ucdavis.edu}
}

\begin{document}
\thispagestyle{plain}
\maketitle
\doublespacing
\begin{abstract}
Manual fruit harvesting is common in agriculture, but the amount of time pickers spend on non-productive activities can make it very inefficient. Accurately identifying picking vs. non-picking activity is crucial for estimating picker efficiency and optimising labour management and harvest processes. In this study, a practical system was developed to calculate the efficiency of pickers in commercial strawberry harvesting. Instrumented picking carts were developed to record the harvested fruit weight, geolocation, and cart movement in real time. These carts were deployed during the commercial strawberry harvest season in Santa Maria, CA. The collected data was then used to train a CNN-LSTM-based deep neural network to classify a picker's activity into ``Pick" and ``NoPick" classes. Experimental evaluations showed that the CNN-LSTM model showed promising activity recognition performance with an F1 score accuracy of over 0.97. The recognition results were then used to compute picker efficiency and the time required to fill a tray. Analysis of the season-long harvest data showed that the average picker efficiency was 75.07\% with an estimation accuracy of 95.22\%. Furthermore, the average tray fill time was 6.79 minutes with an estimation accuracy of 96.43\%. When integrated into commercial harvesting, the proposed technology can aid growers in monitoring automated worker activity and optimising harvests to reduce non-productive time and enhance overall harvest efficiency. \\

\end{abstract}

\section*{Nomenclature}

\begin{table}[ht]
\RaggedRight
\begin{tabular}{ll}
\hline
\multicolumn{2}{l}{Abbreviations} \\
\hline
CNN & Convolutional Neural Network \\
LSTM &Long Short-Term Memory \\
BiLSTM & Bidirectional Long Short-Term Memory \\
CNN-LSTM & Convolutional Neural Network - Long Short-Term Memory \\
GNSS & Global Navigation Satellite System \\
GPS & Global Positioning System \\
CEP & Circular Error Probable \\
IMU & Inertial Measurement Unit \\
IoT & Internet of Things \\
RTK & Real-Time Kinematic \\
SBAS & Satellite-Based Augmentation System \\
SD & Secure Digital (memory card) \\
DBSCAN & Density-Based Spatial Clustering of Applications with Noise \\
LOOCV & Leave-one-out Cross-Validation \\
IQR & Interquartile Range \\
CI & Confidence Interval \\
TP & True Positive \\
TN & True Negative \\
FP & False Positive \\
FN & False Negative \\
\hline
\end{tabular}
\label{tab:abbreviations}
\end{table}

\begin{table}[ht]
\RaggedRight
\begin{tabular}{lll}
\hline
\multicolumn{3}{l}{Symbols and Variables} \\
\hline
$a_x$ & Acceleration in $x$-axis & m/s$^2$ \\
$a_y$ & Acceleration in $y$-axis & m/s$^2$ \\
$a_z$ & Acceleration in $z$-axis & m/s$^2$ \\
m &Mass & kg \\
Height & & m \\
Latitude &  & degrees \\
Longitude & & degrees \\
GNSS time of week & & ms \\
Velocity & & m/s \\
\hline
\end{tabular}
\label{tab:symbols}
\end{table}

\vspace{2em}

\section{Introduction}

Most fresh-market fruits and vegetables are still harvested manually. However, manual harvesting is highly labor-intensive and costly, and the available farm labor workforce has not been able to meet the grower needs \citep{wallstreetbrat,farmlaborshortagehubbart}. Despite advances in robotic harvesting of fruits and vegetables such as strawberries \citep{xiong2020autonomous}, apples \citep{zhang2024automated}, and sweet peppers \citep{arad2020development}, most harvesting robots are still in the prototyping phase and cannot match or exceed manual labor in terms of reliability, cost, speed, and scalability. Growers continue to depend on manual labor for harvesting. Improved labor management and work efficiency could reduce the dependence on labor. However, current methods for quantifying worker efficiency depend on manual observations, such as counting the number of harvested trays or bags per hour or shift. This approach is not scalable and fails to track the temporal dynamics of picker activities. As a result, it is difficult to track the significant amount of time pickers spend on non-productive tasks, such as transporting full containers, waiting in queues for delivery, and repositioning in the field to continue harvesting. 

 Despite potential practical applications, activity recognition has often been overlooked in agricultural contexts. Current activity recognition methods utilise images or time series data, such as acceleration readings. With advancements in Convolutional Neural Networks (CNNs), the accuracy and robustness of vision-based activity recognition methods have significantly improved. \citet{pal2023video} developed a Mask R-CNN and optical flow-based machine vision approach for human activity classification in agriculture. However, agricultural workers operate in cluttered environments characterised by frequent occlusions and diverse worker motions, posing challenges in identifying pickers, tracking them, and recognising their activities throughout harvest. Therefore, it is more practical to develop systems based on direct measurement by sensors that can capture harvest-related signals, such as harvested mass, picker location, or travel speed. Recently, researchers have developed systems such as instrumented picking carts \citep{anjom2018development}, instrumented picking bags \citep{fei2020estimation}, and wearable sensors \citep{dabrowski2023fruit} to measure real-time harvest data and monitor picker motion. While the picking cart developed by \citet{anjom2018development} and the picking bag developed by \citet{fei2020estimation} record harvest mass, acceleration, and location, they were primarily integrated into harvest assist systems rather than for activity recognition or efficiency evaluation. The wearable sensor developed by \citet{dabrowski2023fruit} was designed to detect bag-emptying events during manual avocado harvesting, but did not identify the picker’s picking activity or estimate picker efficiency.
 
Recognising picking activities during manual harvest is challenging since the picking data is embedded within a complex mix of non-picking activities throughout the harvest period. During manual harvest, each picker is provided with picking carts and trays with clamshells for placing fruits like strawberries or larger bags or buckets for apples or avocados. When a tray or bag is full, the pickers move to a designated collection point, often located at the edge of the field or within the rows. If a picker reaches the end of a row with a partially full container, they typically move to a new unharvested row to resume picking. The harvest data contains data-related activities such as active picking, travelling to a collection station for fruit delivery, returning to rows to restart harvesting, switching rows to continue harvesting, as well as idle time before and after harvest and during the breaks. Furthermore, since pickers manually handle the sensing modules, the collected data gets heavily noisy because of the variation in handling, making the activity recognition and efficiency computation challenging.
 
The main goal of this study was to develop a robust data-driven approach to identify active picking events during manual harvest and estimate and evaluate picker efficiency. To achieve this goal, instrumented picking carts (iCarritos) were developed and deployed during commercial strawberry harvest in California, USA. The picking carts were equipped with two load cells, a Global Navigation Satellite System (GNSS) receiver, and an Inertial Measurement Unit (IMU) to record harvested fruit weight, geolocation, and cart motion data. To identify active picking events, a robust Convolutional Neural Network - Long Short-Term Memory (CNN-LSTM) based deep learning algorithm was developed to classify raw harvest data into picking and non-picking ( ``Pick" and ``NoPick") classes. The picker efficiency was then evaluated as the percentage of time spent actively picking over the entire harvest period and the amount of time spent by pickers to fill up a tray. The following are the main contributions of the paper:
\begin{itemize}
    \item Picker efficiency during manual harvest was estimated and evaluated using instrumented picking carts and a robust deep learning-based activity recognition algorithm.
    \item Extensive harvest data was collected during the commercial harvest, and the dataset with complete annotations for training and evaluating the CNN-LSTM model and efficiency estimation has been made publicly available for further research. The dataset can be accessed through the following link.
\end{itemize}

The remaining sections of this paper are organised as follows. Section \ref{sec:materials and methods} details the data collection process, the proposed CNN-LSTM-based activity recognition algorithm, and the picker efficiency estimation process, and discusses the metrics to evaluate the accuracy of picker activity recognition and efficiency estimation. Section \ref{sec:results and discussions} evaluates the accuracy of the activity recognition and efficiency estimation algorithm, followed by the estimation of picker efficiency in harvest data throughout the season. Finally, Section \ref{sec:conclusions} concludes the paper by summarising the key findings and their implications. Details about the released dataset can be found in the "Dataset Details" section of the Appendix.

\section{Materials and Methods}
\label{sec:materials and methods}
\subsection{Data Collection}
\begin{figure}[ht]
    \centering
    \includegraphics[width=0.65\linewidth]{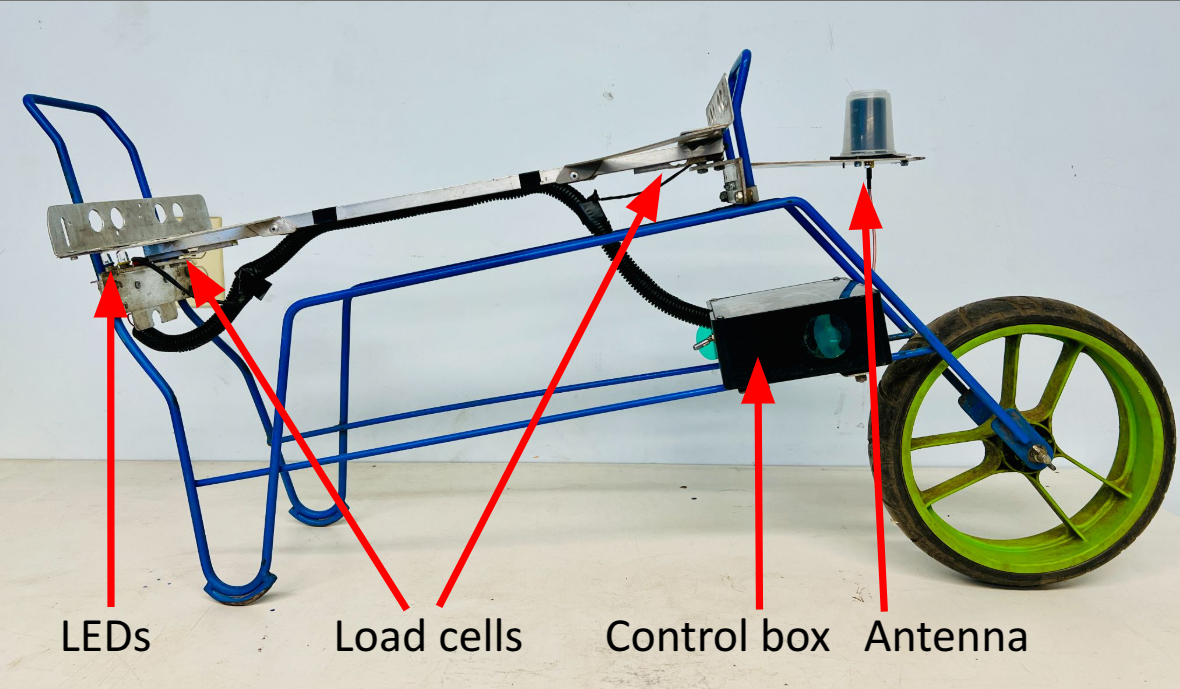}
    \caption{Instrumented picking cart (iCarrito) developed by instrumenting traditional wire frame structure picking carts (Carrito). It consists of two load cells, a SwiftNav Piksi Multi GNSS unit, a control box with a Raspberry Pi 0W microcomputer, an SD card, control switches, and status LEDs. }
    \label{fig:icarrito}
\end{figure}

Data was collected from instrumented picking carts (iCarrito) consisting of two load cells and a GNSS unit with onboard IMU that measured real-time harvest mass, cart location, and motion (see Figure \ref{fig:icarrito}). A Raspberry Pi 0W (Raspberry Pi Foundation, UK) microcomputer with an SD card was used to run the carrito software and store the harvest data. The Piksi Multi (Swift Navigation Inc., USA) GNSS unit was a low-cost GPS that used corrections from the freely available Satellite-Based Augmentation System (SBAS).  The GNSS unit had a rated horizontal Circular Error Probable (CEP) accuracy of 0.75 meters. The harvest data was recorded at 10 Hz with information on Raspberry Pi Unix timestamp, GNSS Unix timestamp, GNSS time of week ($ms$), latitude ($degrees$), longitude ($degrees$), height ($m$), acceleration in x-axis ($m/s^2$), acceleration in y-axis ($m/s^2$), acceleration in z-axis ($m/s^2$), raw mass ($Kg$). Additionally, the centre locations of the strawberry planting beds were collected using RTK GPS to accurately geolocate the harvest row and field boundary.

\begin{figure}[ht]
    \centering
    \includegraphics[width=0.65\linewidth]{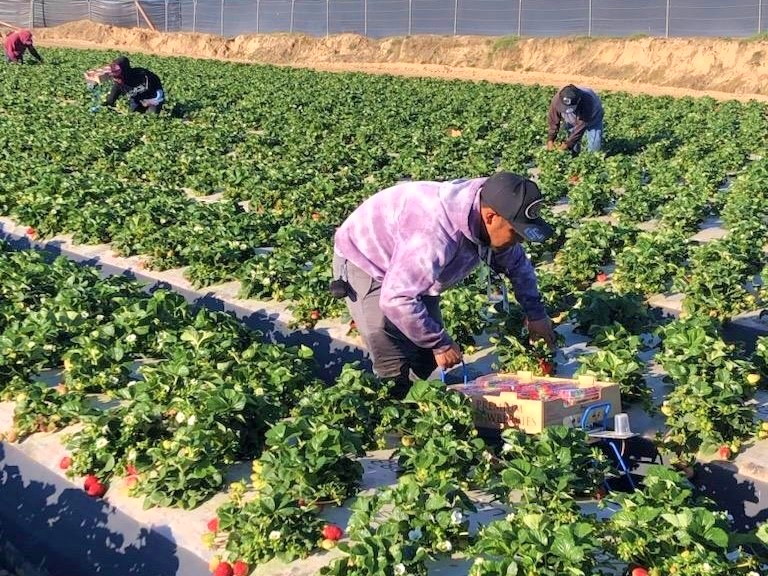}
    \caption{Field deployment of the iCarritos during commercial strawberry harvest in Santa Maria, CA}
    \label{fig:field deployment}
\end{figure}

The iCarritos were deployed during commercial strawberry harvest in Santa Maria, CA (Location: $34^\circ 53' 59.8"" \text{N}, 120^\circ 28' 25.0"" \text{W}$, Area: $17,862.15 m^2$) growing Fronteras strawberry variety in 2024 harvest season. Figure \ref{fig:field deployment} shows the pickers harvesting strawberries with the instrumented picking cart. Each planting bed had a width of 110 cm, with four parallel strawberry plant rows. The harvest season consisted of 27 harvest days (April 10, 2024 - July 17, 2024) with 752 iCarrito deployments. Data collected from early to peak harvest season, covering 12 harvest days (April 10, 2024 - May 25, 2024) were used to train and evaluate the activity recognition model and picker efficiency estimation. Furthermore, data from the entire harvest season, with more than 3000 hours of harvest data, was analysed for a comprehensive analysis of picker efficiency and tray fill time.

\begin{figure}[ht]
    \centering
    \includegraphics[width=0.85\linewidth]{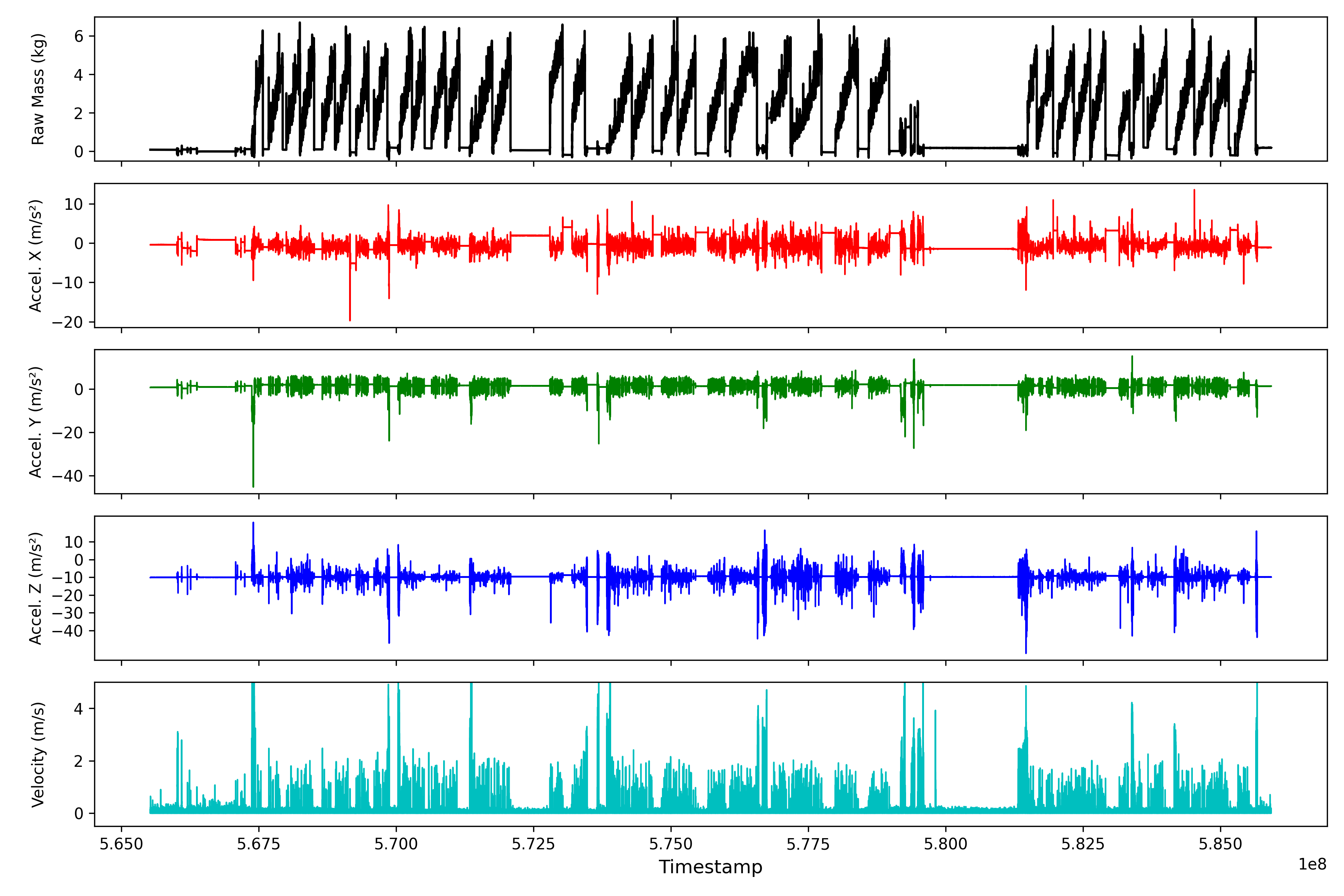}
    \caption{Time series data collected from instrumented strawberry picking carts. The plots show harvested fruit mass; acceleration along the x-axis (red), y-axis (green), z-axis (blue); and velocity (magenta). Each sawtooth structure of the mass data represents full or partially full tray mass increments over time. }
    \label{fig:carrito raw data}
\end{figure}

Figure \ref{fig:carrito raw data} shows the recorded mass and acceleration data by iCarrito during a typical harvest day. The sawtooth-like mass signal indicates an increase in harvest mass as the picker places berries in the tray. Additionally, the periods with almost zero mass represent the times when the pickers were not picking, such as delivering a full tray to the delivery station, returning with an empty tray, switching rows with a partially full tray to continue harvesting, taking meal and personal breaks, and time before and after the harvest. The acceleration and velocity data also show when a picker actively handled their iCarrito, as represented by perturbations along the x, y, and z axes. The velocity data was derived from latitude and longitude measurements. More details on harvest locations and collected data can be found in our other study on precision yield estimation and mapping using instrumented picking carts and a data processing pipeline, which also uses the result of this study for precision yield estimation and mapping \citep{bhattarai2025precision}.

\subsection{Overview of Picker Efficiency Estimation System}
The picker efficiency estimation system consisted of two major steps: CNN-LSTM for activity recognition and the picker efficiency estimation. The objective of the CNN-LSTM was to classify the harvest data into the periods when a picker was actively picking (``Pick" activity) and periods when the picker was not picking (``NoPick" activity).  Once the data was classified into ``Pick" and ``NoPick" classes, the next step was to estimate picker efficiency. However, the ``NoPick" class included all non-picking data, including the data unrelated to harvesting, such as the idle time before the harvest's start, after the harvest's end, and during the meal breaks. Hence, the ``NoPick" data not related to active harvesting were excluded. Finally, the picker efficiency was estimated as the percentage of time spent picking over the entire harvest period and the amount of time required to fill up a tray.

\subsection{CNN-LSTM for Activity Recognition}
\label{sec:cnn-lstm activity rec}
\subsubsection{Data annotation}
The ability of the CNN-LSTM to classify raw data depends on the manually annotated ground truth data used to train the network parameters. In this study, the common approach of manual data annotation was challenging as the recorded data consisted of several million data points. Therefore, the data was annotated using a two-step approach: an automated unsupervised algorithm to label the data, followed by manual refinement with an annotation tool. Algorithm \ref{alg: annotation algorithm} illustrates the unsupervised algorithm to label the raw harvest data as ``Pick" or ``NoPick" label. Non-picking data was identified and removed systematically to design the algorithm. After completing all removal steps, the remaining data was then labelled as picking. In our case, identifying non-picking data was relatively straightforward. The non-picking data was related to recordings outside the field boundary, cases where mass was zero, and occurrences of noisy location jumps that resulted from sensor limitations or from pickers handling the carts. Let the cart data be represented as \( C = \{(x_i, y_i, m_i, a_{x_i}, a_{y_i}, a_{z_i}, t_i)\}_{i=1}^N \). Here, \( (x_i, y_i) \) is the cart's location, \( m_i \) is harvest mass, and \( (a_{x_i}, a_{y_i}, a_{z_i}) \) is cart acceleration along the x, y, and z axes at the corresponding timestamp \( t_i \). The row locations \( R = \{(r_{x_j}, r_{y_j})\}_{j=1}^M \) were measured from the RTK GPS and were used to compute the polygon to outline the field boundary defined by vertex coordinates \( F = \{(x_f, y_f)\}_{f=1}^K \). Let \( C' \) be a modified version of the cart data \( C \), in which non-relevant data points related to non-picking periods were systematically removed.

\begin{algorithm}[ht]
\caption{Data Annotation for Picking Events}
\label{alg: annotation algorithm}
\begin{algorithmic}[1]
\State \textbf{Input:} 

    \Statex \hspace{1em} Cart data ($C$) = $\{(x_i, y_i, m_i, t_i)\}_{i=1}^N$
    \Statex \hspace{1em} Field boundary coordinates ($F$) = $(x_f, y_f)_{f=1}^K$

\State \textbf{Output:} Annotated data \(A\)
\State $C' \gets C$

\State Define field boundary polygon: 
P $\gets$ Polygon(F)
\State Remove data points outside the field boundary: 
\[ C' \gets \{(x_i, y_i, m_i, t_i) \in C | (x_i, y_i) \in P \} \]

\State {Cluster location data and remove noisy location}
    \Statex \hspace{1em} 6.1 $C_{pos} \gets {(x_i, y_i, t_i)}_{i=1}^N$ from $C'$
    \Statex \hspace{1em} 6.2 $clusters \gets \text{DBSCAN}(C_{pos}, \epsilon, minPts)$
    \Statex \hspace{1em} 6.3 $C' \gets \{(x_i, y_i, m_i, t_i) \in C' \mid cluster_{id} \neq noise_{id}\}$
\State Remove noisy weight data: 
 \[C' \gets {(x_i, y_i, m_i, t_i) \in C' \mid m_{\text{min}} < m_i < m_{\text{max}}}\]

 \State Assign picking status to cart data:
\[
A \gets \{(x_i, y_i, m_i, t_i, \text{Pick} \gets 1) \mid (x_i, y_i, m_i, t_i) \in C'\} 
\]
\[
\cup \{(x_i, y_i, m_i, t_i, \text{NoPick} \gets 1) \mid (x_i, y_i, m_i, t_i) \in C \setminus C'\}
\]
\State Return \(A\)
\end{algorithmic}
\end{algorithm}

First, data points located outside the field boundary were removed. To accomplish this, a polygon (P) was created using the field boundary coordinates, and the data points that fell outside the polygon boundary were removed. The noisy location data was then removed using an unsupervised clustering approach. The intuition behind this step was that the data related to the harvested tray was in the form of clusters with closely spaced cluster elements in both time and space. The DBSCAN algorithm was used for data clustering, which assigned the data points to specific clusters or identified them as noise. All data points classified as noise were removed in this step. In the following step, data associated with invalid weight measurements were removed. The weight data less than the empty tray (0.5 Kg) or more than the full tray weight (5.5 Kg) was considered invalid. This step also eliminated data related to idle time, such as when the picker was delivering a full tray and returning with an empty one, time before and after the harvest, during breaks, or an abrupt increase in weight data due to inadvertent pressure applied to the load cells from placing the trays or bumping against the bed edges. In the final step, all the remaining data in the filtered dataset $C'$ was labelled as  ``Pick", and those in the input cart data $C$ but not in filtered data $C'$ were labelled as ``NoPick." Once the data was annotated with the automated algorithm, the annotation was formatted in a format readable by the time-series data annotation tool Label Studio (\url{https://labelstud.io/}). The loaded annotation files were reviewed and updated, if necessary, manually.

\subsubsection{CNN-LSTM architecture}

\begin{figure}[ht]
    \centering
    \includegraphics[width=0.8\linewidth]{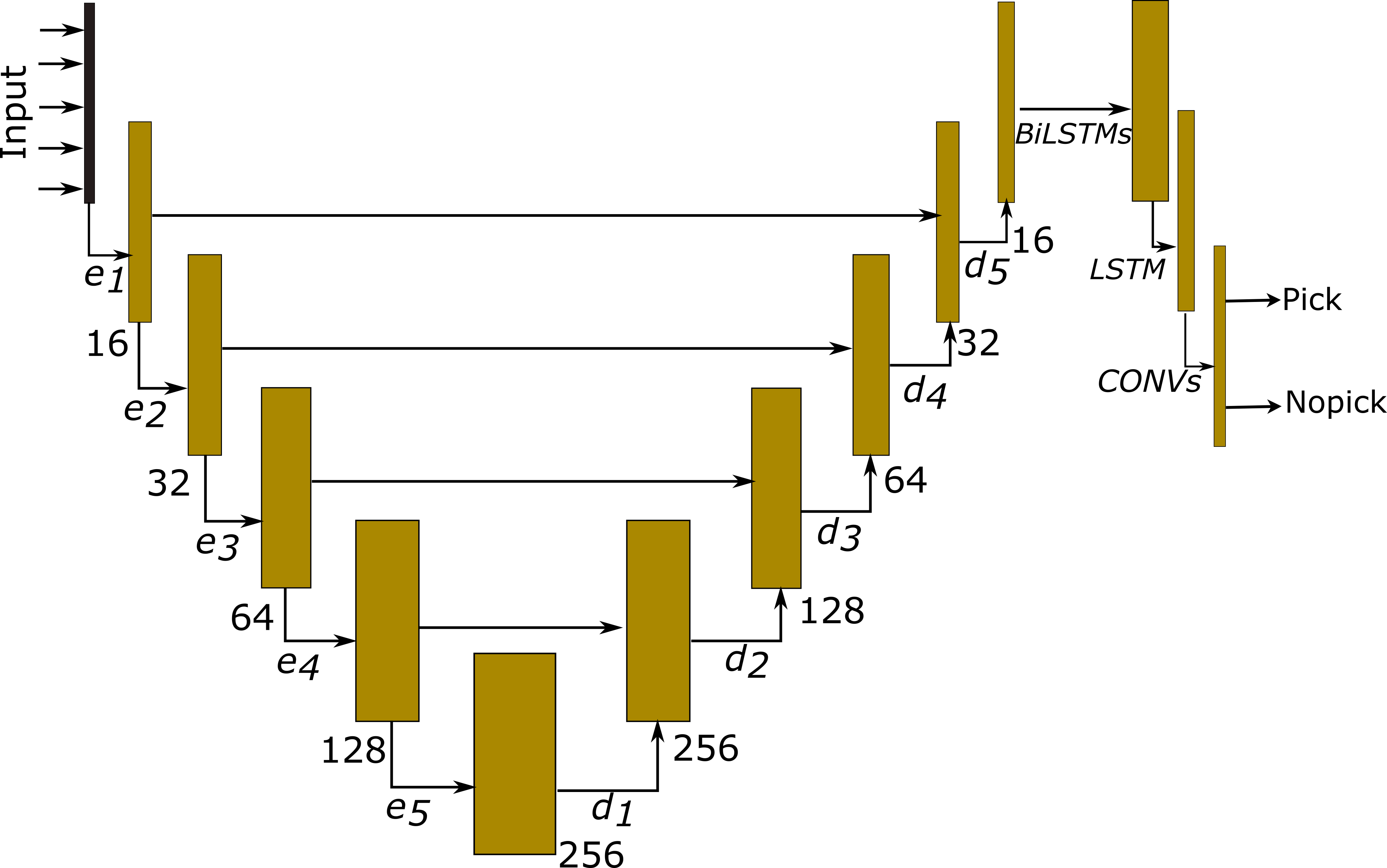}
    \caption{The CNN-LSTM architecture used for picker activity classification. The U-shaped CNN encoder-decoder extracts hierarchical spatial features, while the LSTM layers capture temporal dependencies in the time-series data. The final convolution layers distinguish between ``Pick" and `NoPick" activities."}
    \label{fig:proposed_approach}
\end{figure}

The proposed CNN-LSTM model followed a hybrid architecture with a combination of CNN and LSTM models. Building on the conceptual framework of the CNN model proposed by \citet{perslev2019u}, a novel architecture was developed by integrating the CNN model with LSTM and BiLSTMs while reducing the size and parameters of the CNN model. The model consisted of a U-shaped CNN feature extraction block followed by a combination of BiLSTM and LSTM layers and a classification block (see Figure \ref{fig:proposed_approach}). The U-shaped CNN block extracted the spatial feature from the input data, while the LSTM extracted the long-term time dependencies in the harvest data.  Similar to UNet, the CNN block consisted of encoder and decoder stages with skip connections for hierarchical feature extraction and transmission to the deeper layers. The encoder consisted of five encoding stages ($e_1$ to $e_5$) progressively extracting features from the input data. Each encoder stage included $Conv1d(C_i, C_o, k=9, s=1, p=\text{same})$ layers, followed by Batch Normalisation, ReLU activation, and MaxPooling1D. Here, $C_i$ and $C_o$ were the numbers of input and output channels,  and $k$, $s$, and $p$ were kernel size, step size, and padding, respectively. The number of output feature channels from $e_2$ to $e_5$ in the encoder was increased progressively from 32 to 256 (32, 64, 128, 256) while the feature dimension was decreased by factors of (8, 6, 4, 2).

The decoder section consisted of a series of upsampling and convolution layers to reconstruct the spatial features. The features extracted from the encoders were connected to the decoders at the same feature extraction level through skip connections, providing low-level features to be transmitted into the deeper layers. Each decoding stage started with a feature concatenation, upsampling, convolution [$Conv1d(C_i, C_o, k=9, s=1, p=\text{same})$], BatchNormalization, and ReLU activation. The feature map sizes were increased by the UpSampling1D operations using the same scaling factors as the encoder's MaxPooling1D, but in reverse order: (2, 4, 6, 8). This resulted in a progressive increase in feature dimensions and a decrease in the number of channels through the decoder blocks.

After the data was processed through the CNN block, the resulting feature map was fed to the LSTM block, consisting of two BiLSTM layers and one LSTM layer. The BiLSTM network comprised 64 and 32 units to capture past and future temporal contexts. The final LSTM layer was unidirectional, with 16 units. After the LSTM block, the processed data was passed to the classification block. This classification block consisted of two 1D convolutional layers. The first layer included a hyperbolic tangent activation function, followed by a second layer with a softmax activation function for classification.

\subsubsection{Network training details}

The network was trained up to 50 epochs with a batch size of 270 using Google Colab in a GPU-accelerated environment. The Adam optimiser was employed to optimise the model parameters,  with an initial learning rate of \(1 \times 10^{-3}\) and standard beta values (\(\beta_1 = 0.9\), \(\beta_2 = 0.999\)). Furthermore, our multi-class classification task used the categorical cross-entropy as a loss function. The network was trained using a Leave-one-out Cross-Validation (LOOCV) approach in which data from one harvest day was excluded from training and reserved for testing. The remaining data was split into training and validation sets in an 80:20 ratio. Furthermore, to evaluate the impact of different input features on the classification accuracy of the network, the network was trained using five combinations of input features: i) velocity only; ii) acceleration only; iii) mass only; iv) a combination of mass and acceleration; and v) a combination of mass, acceleration, and velocity.

\subsection{Picker Efficiency Estimation}
\label{sec:predicted efficiency estimation}
The picker efficiency was estimated based on the classification results from the CNN-LSTM model. First, the ``NoPick" data not related to harvesting were removed. The deleted data were those related to before and after the last ``Pick" activity and the duration of breaks. The number of breaks during harvest was manually identified from the raw data. The pickers took breaks simultaneously, and these breaks could be recognised as an extended period when most of the carts were idle. After removing data unrelated to harvesting, picker efficiency was calculated by estimating the percentage of time spent actively picking over the harvest period. The picker efficiency was calculated as:
\begin{equation}
    \text{Picker Efficiency} = \frac{\text{Total pick time}}{\text{Total harvest time}} \times 100\%
    \label{eq:harvesting percentage}
\end{equation}
The ``Total harvest time" was the valid harvest time after excluding the idle periods before and after the harvest and during breaks, while the ``Total picking time" was the cumulative time spent actively picking. In addition to picker efficiency, the average tray fill time was calculated based on the amount of time required to fill a tray:

\begin{equation}
    \text{Tray fill time} = \frac{\text{Total pick time}}{\text{T}}
    \label{eq:tray fill time}
\end{equation}
``T" was the total number of harvested trays by a cart. The number of harvested trays was manually counted from the raw harvest data and validated by cross-referencing with the harvest tray counts provided by the grower to compensate the pickers.

\subsubsection{Picker Efficiency Estimation from Season-long Harvest Data}
The picker efficiency for season-long harvest (April 10, 2024 - July 17, 2024) was estimated by processing the entire season harvest data collected from 778 iCarritos, recording more than 3000 hours of harvest data. The activity recognition algorithm first classified all the data, followed by picker efficiency and tray fill time estimation.  The Interquartile Range (IQR) method was used to identify and remove outliers from the harvest data by setting the outlier thresholds at 1.5 times the IQR below the first quartile and above the third quartile.

\subsection{System Performance Evaluation}
\label{sec:performance evaluation}

\subsubsection{Evaluation of CNN-LSTM for activity recognition}
The model performance was evaluated using a five-fold Leave-one-out Cross-Validation (LOOCV) approach. Five separate experiments were conducted, where data from one harvest day were reserved for testing. The classification output for each experiment was then compared against the manually annotated ground truth by computing Precision, Recall, and F1 score metrics. The final model performance was then calculated as the average of these metrics across five experiments.

\subsubsection{Evaluation of picker efficiency estimation}
To evaluate the picker efficiency estimation algorithms, the ground truth for the picker efficiency data was computed from the ground truth annotations used to train and evaluate the CNN-LSTM network. The data corresponding to idle time before and after the harvest and rest breaks were removed from the annotated data. The ground truth picker efficiency and tray fill time were estimated as illustrated in equations \ref{eq:harvesting percentage} and \ref{eq:tray fill time}. Finally, the estimated picker efficiency and tray fill time were evaluated as given below.
\begin{equation}
\text{Picker efficiency estimation accuracy} = \left(1 - \frac{1}{C} \sum_{i=1}^C \frac{|\text{gt\_efficiency}_{i} - \text{est\_efficiency}_{i}|}{\text{gt\_efficiency}_{i}}\right) \times 100\%
\end{equation}

\begin{equation}
\text{Tray fill time estimation accuracy} = \left(1 - \frac{1}{C} \sum_{i=1}^C \frac{|\text{gt\_trayfilltime}_{i} - \text{est\_trayfilltime}_{i}|}{\text{gt\_trayfilltime}_{i}}\right) \times 100\%
\end{equation}

Where $C$ was the total number of carritos used for evaluation, gt\_efficiency$_i$ and gt\_trayfilltime$_i$ were ground truth efficiencies and tray fill times, respectively, and est\_efficiency$_i$ and est\_trayfilltime$_i$ were estimated values for carrito $i$. 

\section{Results and Discussions}
\label{sec:results and discussions}
\subsection{Evaluation of CNN-LSTM for activity recognition}
Table \ref{tab:performance_metrics} shows the average precision, recall, and F1 score of the CNN-LSTM model trained on different combinations of input training features. All the input training features showed promising results, with precision, recall, and F1 scores of more than 0.97, except for the velocity data. Using velocity as input training data resulted in the lowest precision (0.945), recall (0.953), and F1 score (0.948), indicating that the velocity data alone may not be sufficient as a standalone harvest indicator. This limitation could be because velocity was derived from GPS location data. Since the iCarritos relied on the low-cost GNSS unit with SBAS-based corrections, the localisation accuracy may not have been sufficient to compute precise cart velocity to capture harvest-related signals. Training on acceleration data demonstrated competitive performance with a precision of 0.963 and an F1 score of 0.972, while training using only mass data achieved even better results with the highest recall (0.984) and F1 score (0.974). The combination of acceleration and mass data also showed excellent performance with high average precision (0.964), recall (0.984), and F1 scores (0.973). Finally, the combination of mass, IMU, and velocity achieved a high precision (0.966) but a low recall score (0.979), resulting in a reduced F1 score of 0.972. 

\begin{table}[ht]
\caption{Performance metrics of the CNN-LSTM for different training feature sets}
\centering
\begin{tabular}{lccc}
\hline
Training features & Precision & Recall & F1 \\
\hline \hline
\rule{0pt}{2.5ex}
velocity & 0.945 & 0.953 & 0.948 \\
\rule{0pt}{2.5ex}
ax, ay, az & 0.963 & 0.983 & 0.972 \\
\rule{0pt}{2.5ex}
mass & 0.965 & \textbf{0.984} & \textbf{0.974} \\
\rule{0pt}{2.5ex}
mass, ax, ay, az  & 0.964 & \textbf{0.984} & 0.973 \\
\rule{0pt}{2.5ex}
mass, ax, ay, az, velocity & \textbf{0.966} & 0.979 & 0.972 \\
\hline
\end{tabular}

\label{tab:performance_metrics}
\end{table}

\begin{figure}[ht]
    \centering
    \includegraphics[width=0.95\linewidth]{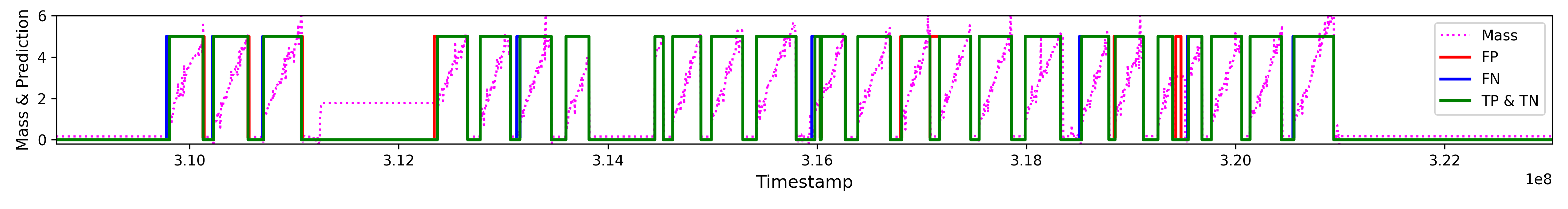}
    \includegraphics[width=0.95\linewidth]{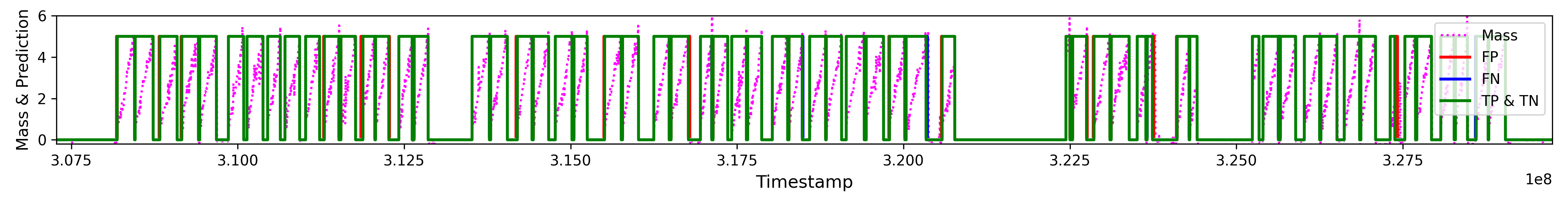}
    \caption{Classification results of picker activity recognition during early (April 24, 2024, top) and peak (May 25, 2024, bottom) periods of harvest season. The square wave-like structures represent the model predictions. True positives (TP) and true negatives (TN) are marked in green, false positives (FP) in red, and false negatives (FN) in blue. The magenta colour indicates the harvest mass over time.}
    \label{fig:classification results}
\end{figure}

The results suggest that the mass and acceleration measurements captured weight change and the cart motion pattern, which is distinctive to the picking and non-picking events. While training only with mass data showed better results, combining acceleration with mass could provide more robust results in unseen data because both sensors capture the unique harvesting contexts. Therefore, for the purpose of picker efficiency computation, the model trained on both mass and acceleration data was used to balance prediction accuracy and generalizability in real-world harvesting conditions. The decrease in F1 score while combining mass, acceleration, and velocity was likely due to network overfitting and potential noise from GPS errors. The network may overfit because the feature information from velocity data may also be extracted from the IMU data, resulting in redundant training features. Figure \ref{fig:classification results} visually represents the classification results during the season's early and peak harvest period. The upper plot shows harvest data from the beginning of the season, characterised by a higher tray-filling time (low mass slope with time), with only a few boxes harvested per day. In contrast, the lower plot represents the peak harvest season with substantially lower tray-filling times (high mass slope) and larger numbers of filled trays within the same harvest period. In Figure \ref{fig:classification results}, the predicted true positives (TP) and true negatives (TN) are marked in green, while false positives (FP) and false negatives (FN) are marked in red and blue, respectively. Despite the substantial differences in tray-filling time and mass data slope between these two season phases, the developed network demonstrated robust, consistent, and promising performance.

\subsection{Evaluation of picker efficiency estimation}

\begin{table}[ht]
\centering
\caption{Statistical summary of the accuracy of estimating picker efficiency and tray fill time}
\begin{tabular}{lcc}
\hline
Statistic & Picker efficiency (\%) & Tray fill time (\%) \\
\hline \hline
Mean & 95.22 & 96.43 \\
Median & 96.58 & 97.60 \\
Range & 78.65 -- 99.93 & 77.87 -- 99.93 \\
Standard Deviation & 4.22 & 3.87 \\
\hline
\end{tabular}
\label{tab:accuracy_stats}
\end{table}

\begin{figure}[ht]
    \centering
    \includegraphics[width=0.6\linewidth]{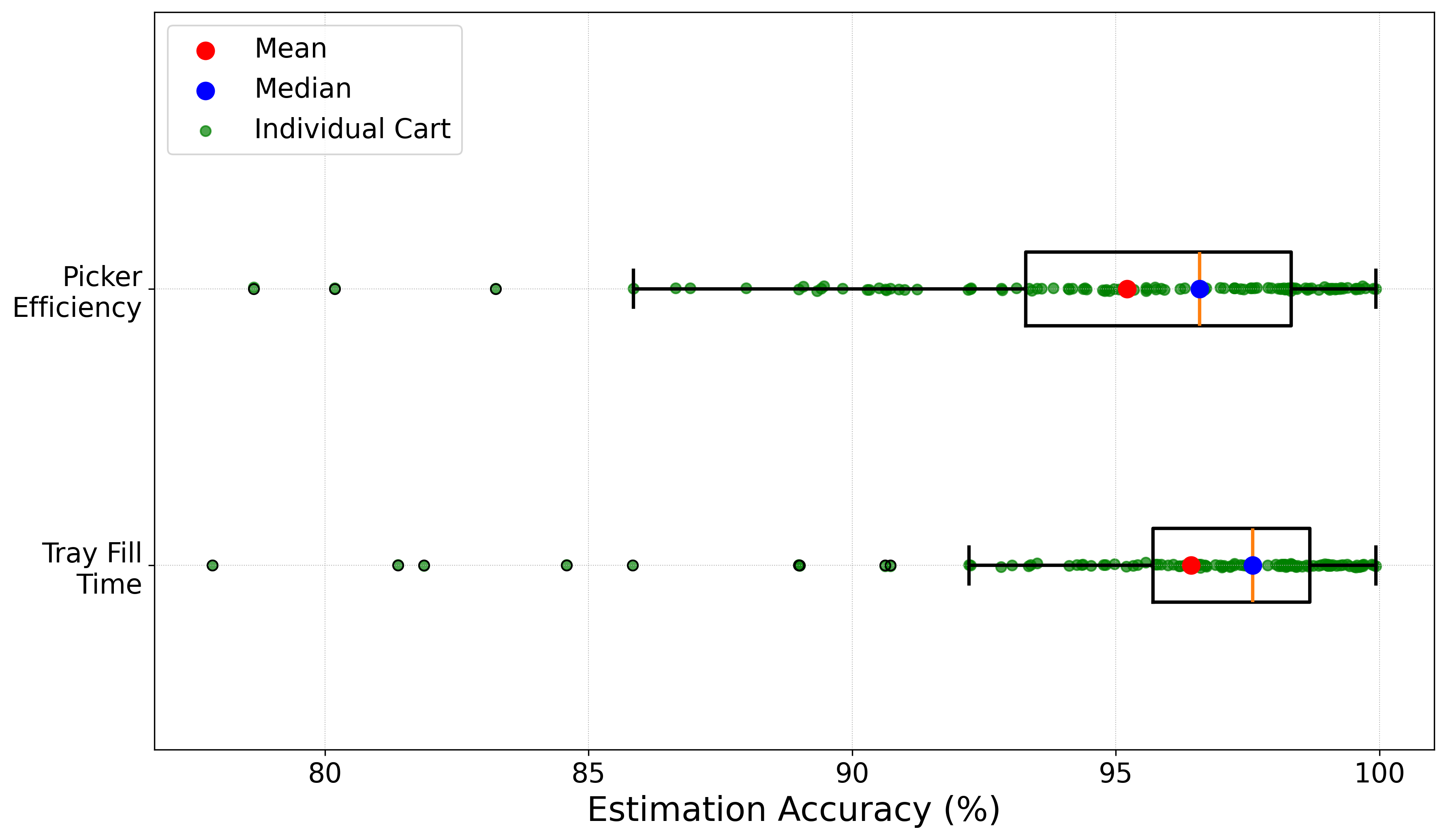}
    \caption{Box plot showing the accuracy of picker efficiency and tray fill time estimation. The mean and median estimation accuracy of both metrics was consistently above 95\%.}
    \label{fig:box plot estimation accuracy}
\end{figure}

The picker efficiency estimation also demonstrated impressive results, with a mean and median estimation accuracy of over 95\% and 96\% for both the picker efficiency and tray fill time. Table \ref{tab:accuracy_stats} presents the statistical summary, and Figure \ref{fig:box plot estimation accuracy} provides the box plot representations of both estimation accuracies. The results showed that the developed algorithms were robust in estimating picker efficiency throughout the observed harvest period. While the mean and median accuracies were high, there were a few outliers, and the results may be influenced by factors such as dataset anomalies, including sensor glitches, unusual cart movements, or the limitations of the neural network model. Nevertheless, the low standard deviation (3.87 for tray fill time and 4.22 for picking efficiency) indicates that estimation accuracy clustered closely around the mean with reliable results.

\subsection{Estimation of Picker Efficiency from Season-long Harvest Data}
\begin{figure}[ht]
    \centering
    \includegraphics[width=0.48\linewidth]{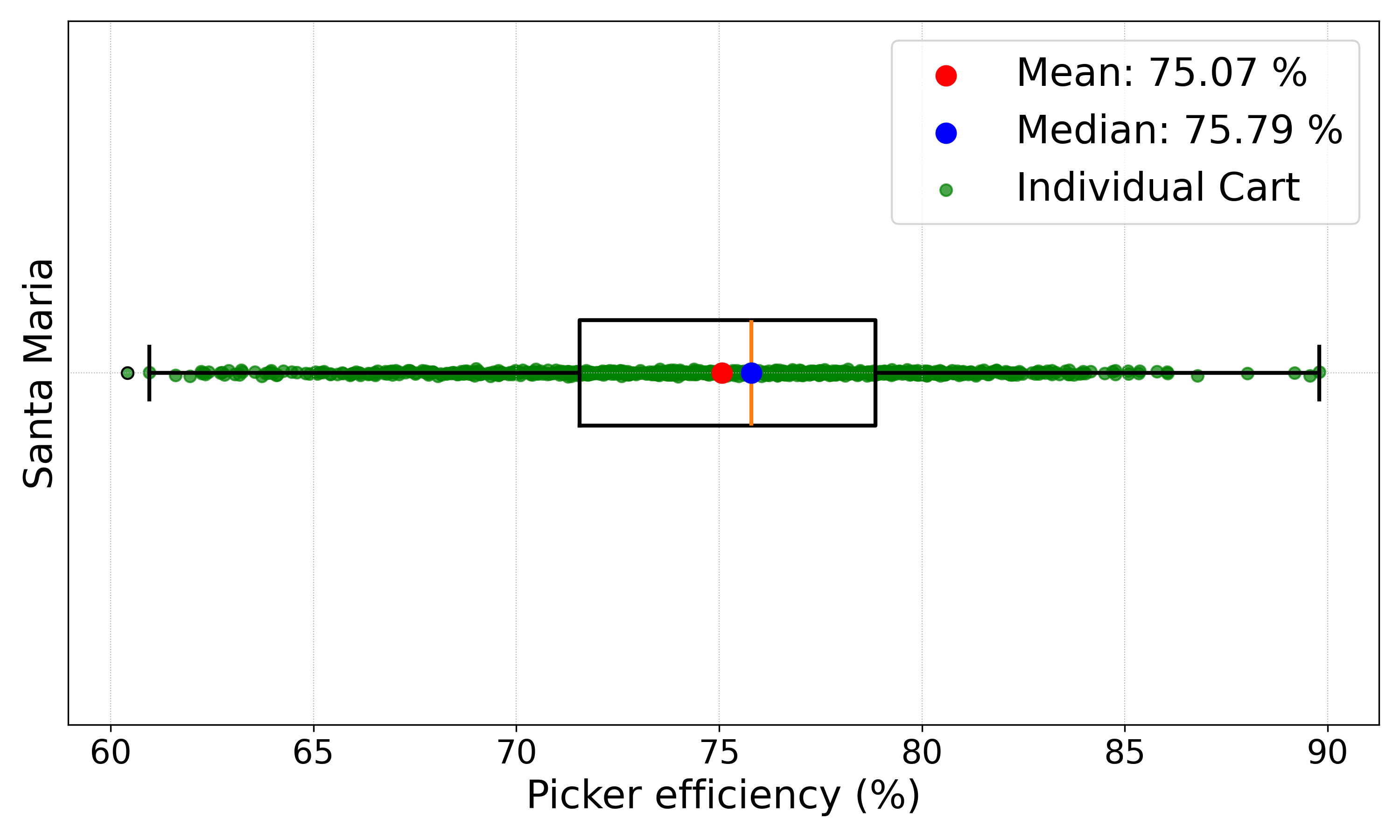}
    \includegraphics[width=0.48\linewidth]{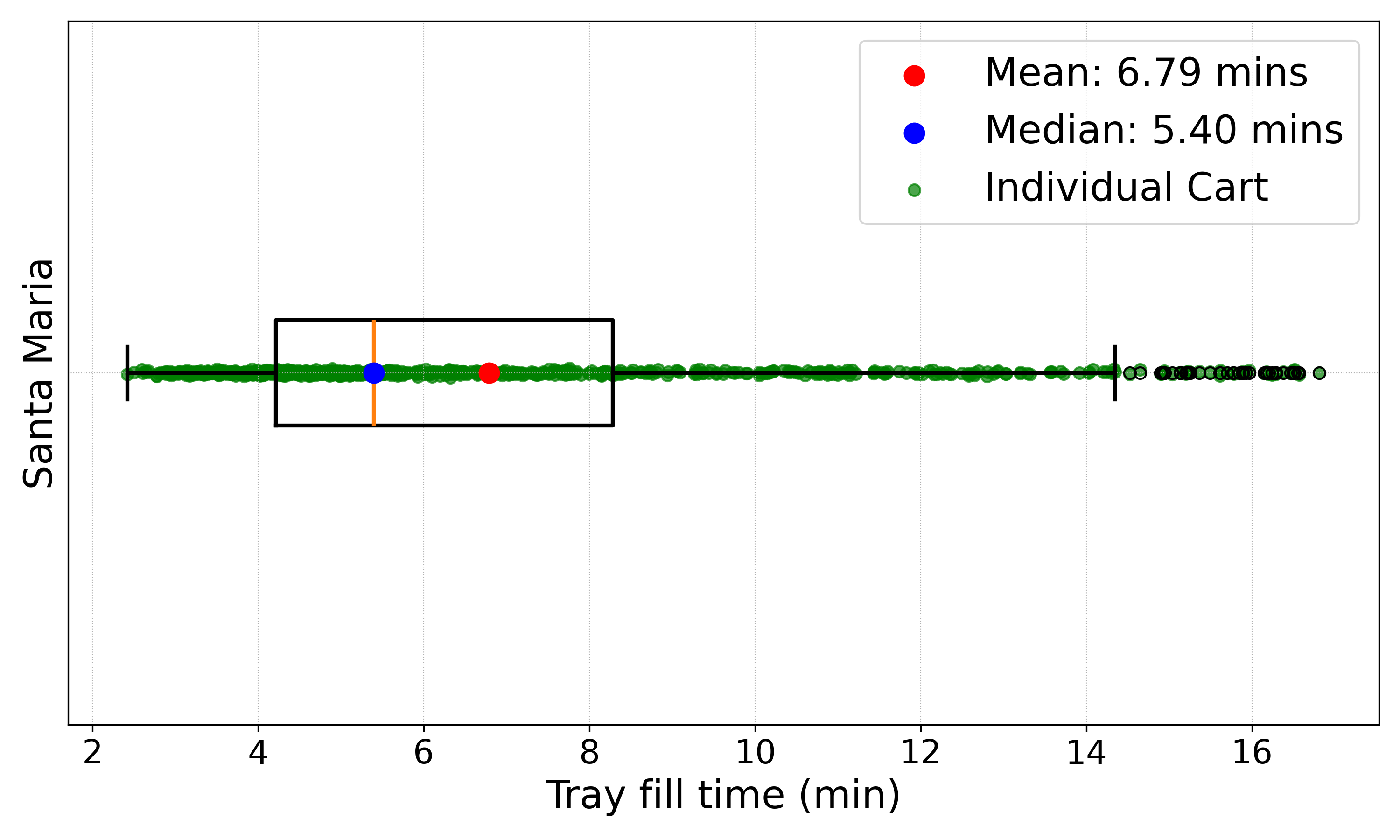}
    \includegraphics[width=0.48\linewidth]{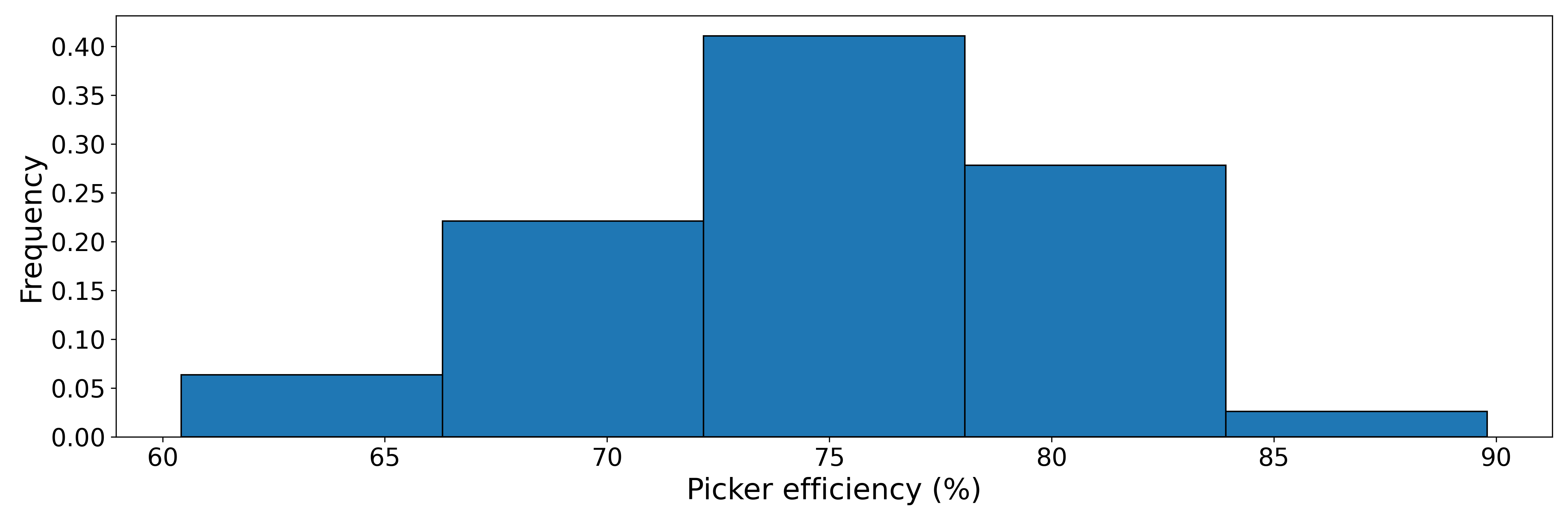}
    \includegraphics[width=0.48\linewidth]{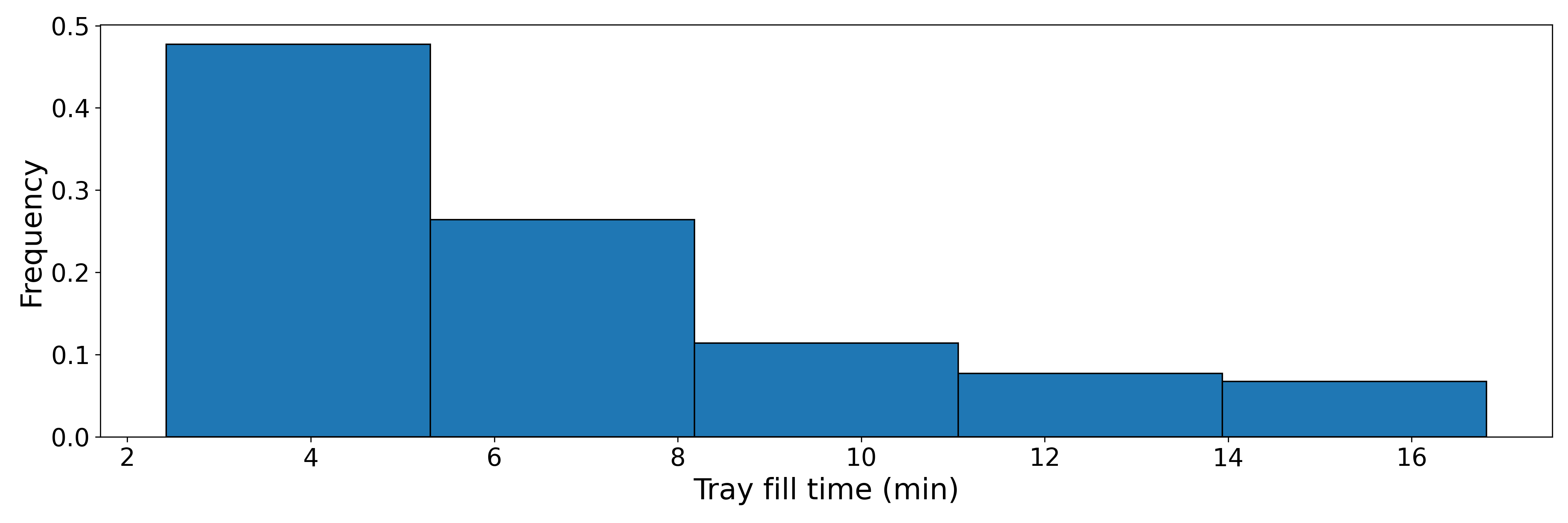}
    \includegraphics[width=0.48\linewidth]{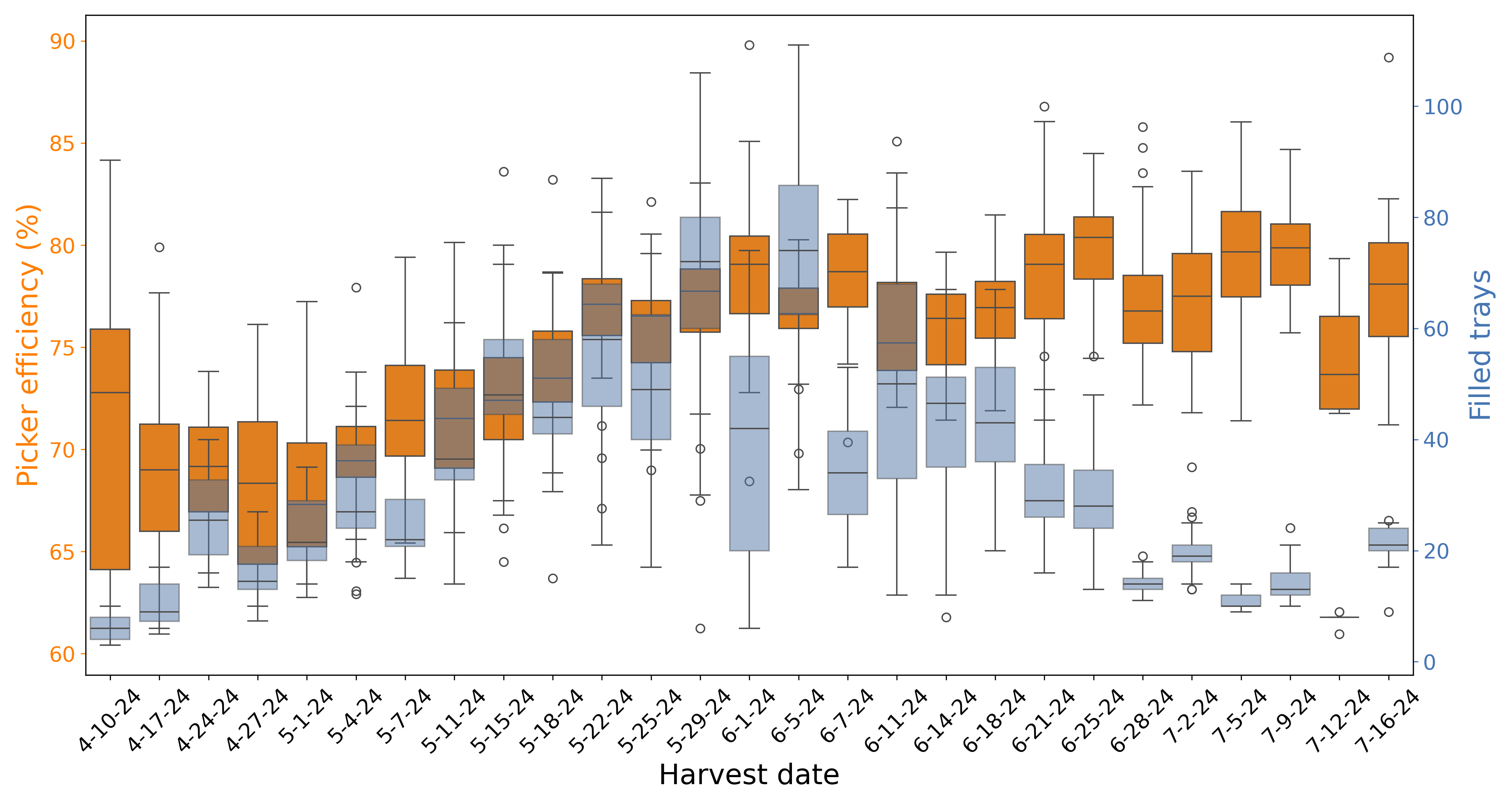}
    \includegraphics[width=0.48\linewidth]{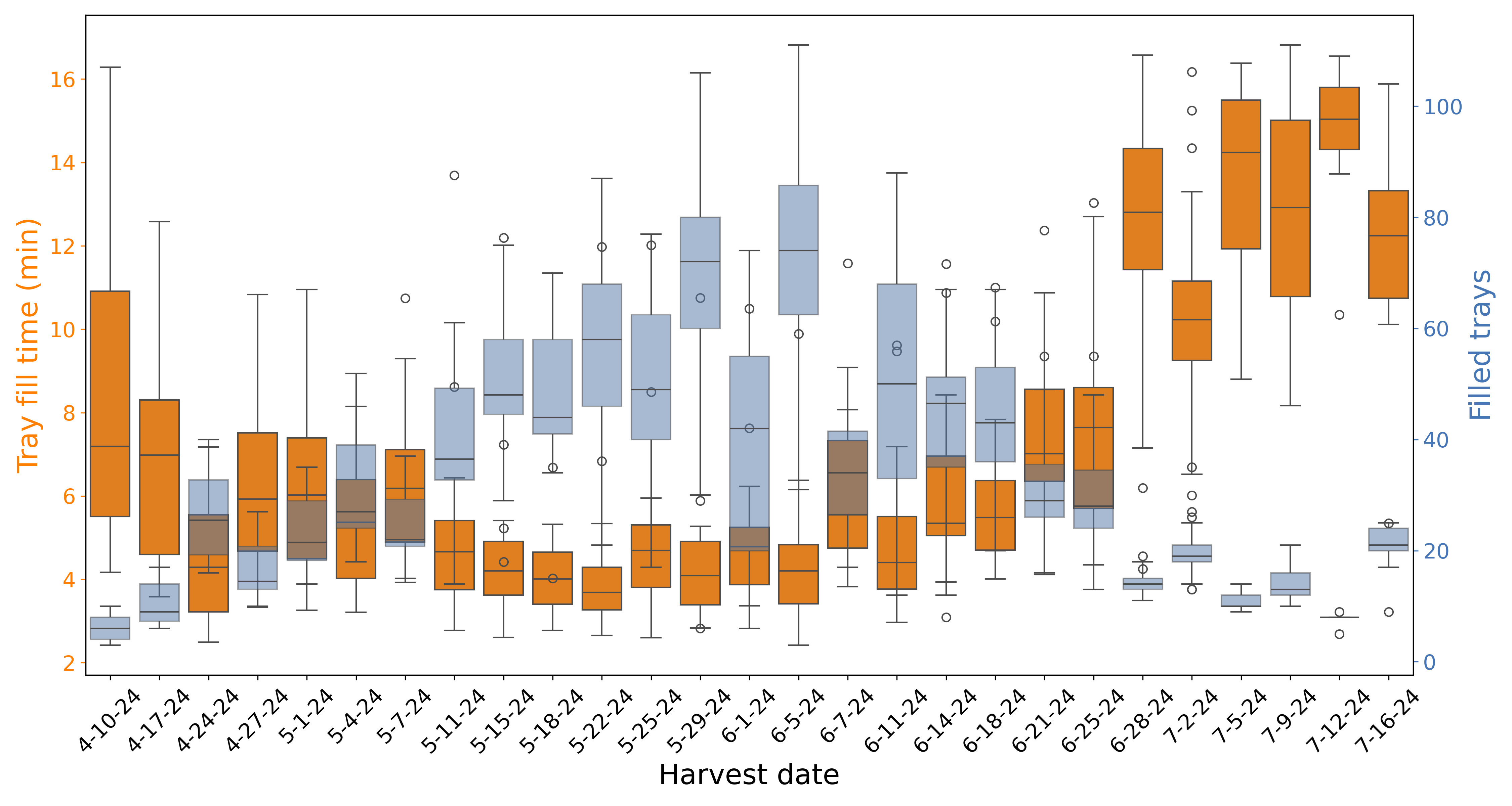}
    \caption{Seasonal picker efficiency analysis: a comprehensive analysis of picker efficiency throughout the strawberry harvest season in Santa Maria, CA. The top and middle rows show the box plots and frequency distribution of the picker efficiency (left) and tray fill time (right). The bottom row shows the trend of the picker efficiency (left), tray fill time (right), and the number of filled trays per cart from the beginning to the end of the harvest season.}
    \label{fig:seasonal_picker_efficiency}
\end{figure}

The picker efficiency analysis on season-long harvest data showed that, on average, pickers spent 75.07\% (95\% CI: [74.70, 75.44]) of their total harvest time actively picking strawberries.  This suggests that pickers spent approximately 24.93\% of the total harvest time on non-picking activities such as travelling to and from the headland to deliver full trays and retrieve empty ones, waiting in delivery queues, switching between rows to continue harvesting, and taking personal or restroom breaks. Furthermore, pickers spent an average of 6.79 minutes (95\% CI: [6.54, 7.04]) filling a tray, with individual tray fill times ranging from 2.42 minutes to 16.81 minutes. Table \ref{tab:seasonal_picker_efficiency} presents the statistical summary of estimated picker efficiency metrics, and Figure \ref{fig:seasonal_picker_efficiency} shows the box plots and frequency distributions of the picker efficiency and tray fill time over the harvest season. The frequency distribution plot (Figure \ref{fig:seasonal_picker_efficiency}, centre) shows that approximately 40\% of picker efficiency was around 75\%, while approximately 60\% of trays were filled within 6 minutes.

\begin{table}[h!]
\centering
\caption{Summary of picker efficiency estimation for season-long harvest}
\begin{tabular}{lcc}
\hline
Statistic & Picker efficiency (\%) & Tray fill time (mins) \\
\hline \hline
Mean & 75.07 & 6.79 \\
Median & 75.79 & 5.40 \\
Range &60.42 -- 89.80 &2.42 -- 16.81 \\
Standard Deviation & 5.23 & 3.50 \\
95\% CI & [74.70, 75.44] & [6.54, 7.04] \\
\hline
\end{tabular}
\label{tab:seasonal_picker_efficiency}
\end{table}

The location of the collection station may play a larger role in explaining the variability of individual picker efficiency than the area or yield of the field. Collection stations are positioned at headlands, and different pickers may need to travel different distances to deliver full trays, even though the yield was similar. Furthermore, during the early and late parts of the season, when yields were lower, pickers spent less time travelling to the collection station and more time actively harvesting, influencing the variability in picker efficiency. On the other hand, yield may have completely influenced the tray fill time. Higher yield results in a faster tray fill rate and vice versa.  The seasonal trend (Figure \ref{fig:seasonal_picker_efficiency}, bottom right) indicates that tray fill times were higher at the beginning of the season due to lower yields. As the season progressed and the yields increased, the tray fill times decreased, only to increase again toward the end of the harvest season as yields diminished.

The outcome of this study, obtained by analysing more than 3,000 hours of harvest data, presents a novel approach to estimate the current standards of manual fruit harvesting. Furthermore, the proposed approach's activity recognition and efficiency estimation results have promising potential for practical applications in agriculture. The information on individual picker efficiency could be utilised to refine labour management strategies. Furthermore, methods to optimise collection station locations can be implemented based on information on the non-productive time to minimise delays and streamline the picker workflow. Beyond labour management, these results could also help in broader field management strategies. Information on tray fill time across different field locations helps growers identify yield variations. Additionally, it can help optimise the scheduling of harvest assistance systems, such as crop transportation systems, for transporting filled trays from the field to collection stations, improving harvest efficiency. The findings of this study have already been implemented for precision yield estimation and mapping in our other research \citep{bhattarai2025precision}.

\section{Conclusions}
\label{sec:conclusions}
In this study, an automated picker activity recognition and efficiency estimation system was developed for manual strawberry harvest using instrumented picking carts (iCarritos) and a CNN-LSTM deep learning model. Harvest data was collected using iCarritos during commercial strawberry harvests, followed by activity recognition and efficiency estimation. Based on the results of this study, the following conclusions were drawn.
\begin{itemize}
    \item CNN-LSTM could provide robust performance for activity recognition tasks in agriculture using time series mass and acceleration data, achieving an F1 score of more than 0.97 in classifying picker activities into ``Pick" and ``No Pick" categories.
    
    \item The results from the CNN-LSTM model could be post-processed to reliably estimate picker efficiency, with a mean active harvesting percentage estimation accuracy of 95.22\% and tray fill time estimation accuracy of 96.43\%.
    \item Analysis of the season-long harvest data revealed that pickers spent, on average, 75.07\% of their total harvest time actively picking strawberries, with the remaining 24.56\% attributed to non-picking activities. The average tray fill time was found to be 6.79 minutes.
\end{itemize}
The developed system provides a practical solution for monitoring automated worker activity in commercial strawberry fields. By providing detailed data on picker efficiency, growers can identify areas for improvement, optimise resource allocation, and enhance overall harvest efficiency. The publicly released annotated dataset further contributes to advancing research in this area. 

\subsubsection*{Acknowledgments}
This work was funded by the USDA Agricultural Research Service (USDA-ARS) Areawide Pest Management Grant Program project ``Site-Specific Soil Pest Management in Strawberry and Vegetable Cropping Systems Using Crop Rotation and a Needs-Based Variable Rate Fumigation Strategy” through Non-Assistance Cooperative Agreements 58-2038-9-016 and 58-2038-3-029. We would also like to acknowledge the support of Dennis Lee Sadowski and Alejandro Torres Orozco, whose contributions are invaluable to the success of this study.

\bibliographystyle{apalike}

\section*{Appendix: Dataset Details}
\label{appendix}

The publicly released dataset consists of two main folders, \texttt{cart\_recorded\_data} and \texttt{break\_log}, as well as an Excel file named \texttt{harvested\_trays.xlsx}. The following section provides detailed description of the structure and contents of each component:

\begin{enumerate}
    \item \textbf{\texttt{cart\_recorded\_data} folder:}
    This folder contains all annotated harvest data collected during the study.
    It includes twelve CSV files, each corresponding to a specific harvest day. The file naming convention is \texttt{mm-dd-yy\_train-ready\_all\_carts.csv} (e.g., \texttt{4-10-24\_train-ready\_all\_carts.csv}). The date in the filename indicates the harvest date. The suffix \texttt{train-ready\_all\_carts} signifies that the data is preprocessed and ready for training and evaluation.Each CSV file contains nine columns, described as follows:
        \begin{itemize}
            \item \texttt{date\_cartID}: Combined harvest date and cart identifier.
            \item \texttt{GPS\_TOW}: GPS Time of Week (milliseconds).
            \item \texttt{easting}, \texttt{northing}: UTM coordinates (meters).
            \item \texttt{ax}, \texttt{ay}, \texttt{az}: Acceleration along the $x$, $y$, and $z$ axes (m/s$^2$).
            \item \texttt{raw\_mass}: Raw harvest mass (kg).
            \item \texttt{activity}: Annotated activity label, indicating whether the data point corresponds to a ``Pick'' or ``NoPick'' event.
        \end{itemize}

    \item \textbf{\texttt{break\_log} folder:}
    This folder contains break-related data for all carts included in \texttt{cart\_recorded\_data}. It consists of twelve CSV files, each named according to the format \texttt{mm-dd-yy\_break\_log.csv} (e.g., \texttt{4-10-24\_break\_log.csv}). Each CSV file contains the following columns:
        \begin{itemize}
            \item \texttt{harvest\_date}: Harvest date.
            \item \texttt{\#carrito}: Cart identifier.
            \item \texttt{no\_breaks}: Number of breaks taken by the corresponding cart on that day.
        \end{itemize}

    \item \textbf{\texttt{harvested\_trays.xlsx} file:}
    This Excel file records the number of trays harvested by each cart (\textit{carrito}) for each harvest day. Data for each harvest day is stored in a separate worksheet (tab). Each worksheet contains two columns:
        \begin{itemize}
            \item \texttt{\#carrito}: Cart identifier.
            \item \texttt{\#trays\_carrito}: Number of trays harvested by the corresponding cart on that day.
        \end{itemize}
\end{enumerate}

The dataset provides comprehensive, annotated time-series data suitable for training and evaluating activity recognition models, as well as for further research in harvest process optimization and labor efficiency analysis.

\end{document}